\documentclass[10pt, a4paper]{article}

\usepackage{lrec-coling2024} 
\usepackage{multirow} 
\usepackage{url}
\title{LexDrafter: Terminology Drafting for Legislative Documents using Retrieval Augmented Generation}

\name{Ashish Chouhan and Michael Gertz} 

\address{Institute of Computer Science, Heidelberg University \\
	Im Neuenheimer Feld 205, Heidelberg, Germany \\
	\{chouhan, gertz\}@informatik.uni-heidelberg.de\\}

\abstract{
With the increase in legislative documents at the EU, the number of new terms and their definitions is increasing as well. As per the Joint Practical Guide of the European Parliament, the Council and the Commission, terms used in legal documents shall be consistent, and identical concepts shall be expressed without departing from their meaning in ordinary, legal, or technical language. Thus, while drafting a new legislative document, having a framework that provides insights about existing definitions and helps define new terms based on a document's context will support such harmonized legal definitions across different regulations and thus avoid ambiguities. In this paper, we present \textit{LexDrafter}, a framework that assists in drafting \emph{Definitions} articles for legislative documents using retrieval augmented generation (RAG) and existing term definitions present in different legislative documents. For this, definition elements are built by extracting definitions from existing documents. Using definition elements and RAG, a \emph{Definitions} article can be suggested on demand for a legislative document that is being drafted. We demonstrate and evaluate the functionality of \textit{LexDrafter} using a collection of EU documents from the energy domain. The code for \textit{LexDrafter} framework is available at
\url{https://github.com/achouhan93/LexDrafter}.
 \\ \newline  }

\begin{document}

\maketitleabstract

\section{Introduction}
Around 24k regulatory documents are accessible via the EUR-Lex platform\footnote{\url{https://eur-lex.europa.eu/statistics/2022/eu-law-statistics.html?locale=en}, accessed on 11th September 2023} for the year 2022. 15{\%} of these documents are legislative documents (also called legal acts), comprising \textit{regulations}, \textit{directives}, \textit{decisions}, \textit{recommendations}, and \textit{opinions}. 

Legal acts are considered a vital resource for protecting and safeguarding individuals and organizations. However, the domain-specific terminologies used in these documents are complex, thus complicating the process of reviewing texts and leading to different interpretations of the same content by different users~\cite{Damaratskaya2023}. Similarly, drafting a legal document requires manual effort and domain knowledge to capture the relevant context and details for the document~\cite{DBLP:conf/icail/LamCY23}. As legal documents are usually highly structured and standardized, the manual steps involved in document drafting are time-consuming, resource-intensive, and prone to human error, resulting in a significant potential to automate the document drafting process~\cite{DBLP:journals/corr/abs-2109-11603}.

\begin{figure}[htbp]
	\centering
	\includegraphics[width=0.5\textwidth]{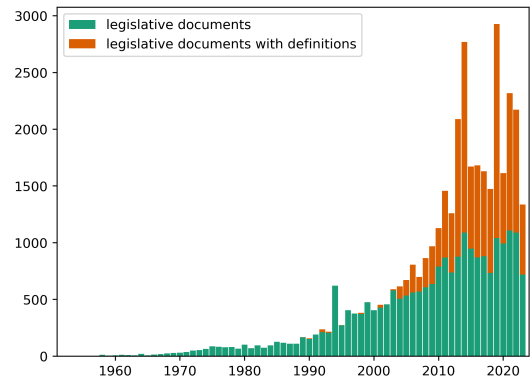}
	\caption{Number of legal documents (with definitions) per year.} 
	\label{fig:yeardistribution}
	\vspace*{-2ex}
\end{figure}

~\citet{gauci2016vessel} pointed out that during document drafting, there are situations where different legal documents assign different definitions to a legal term. In general, having harmonized legal definitions is an essential part of legal drafting practice for precise and effective communication~\cite{DBLP:conf/jurix/AmaludinWPP21}. As per the Joint Practical Guide (JPG) of the European Parliament, the Council and the Commission, for people involved in drafting European Union legislation~\cite{doi/10.2880/5575}, guidelines 6 and 14 provide information about drafting of the terminologies mentioned in a document. Considering these guidelines and the steady increase in the number of documents containing definitions, as shown in Figure \ref{fig:yeardistribution}, a framework that extracts or ``generates'' definition elements from existing documents might serve as an important supporting tool in drafting legal documents and definitions sections in particular. This need is further justified by looking at some statistics. For example, on average, 20 definitions are present in legal acts for the domain \textit{Energy}, which is one of 20 domains\footnote{\url{https://eur-lex.europa.eu/browse/directories/legislation.html}, \\accessed on 1st March 2024}. But there are also outliers, e.g., the document on ``establishing a guideline on electricity transmission system\ldots'' (Celex ID 32017R1485) has 159 definitions.


As technology advances, AI-powered tools can assist legal professionals and expedite the writing process, e.g., LegalSifter\footnote{\url{https://www.legalsifter.com}, \\accessed on 24th August 2023}, Kira Systems\footnote{\url{https://www.kirasystems.com}, \\accessed on 24th August 2023}, or LawGeex\footnote{\url{https://www.lawgeex.com}, \\accessed on 24th August 2023} are AI-powered tools that assist in drafting contracts, identifying and extracting information from contracts, and perform contract analysis. With the advent of Chatbots, legal services platform such as LawDroid\footnote{\url{https://lawdroid.com/}, \\accessed on 24th August 2023} or Law ChatGPT\footnote{\url{https://lawchatgpt.com/}, \\accessed on 24th August 2023} use templates and questionnaires to assist drafting contracts, wills, or agreements. Another product, Lexis+ AI\footnote{\url{https://www.lexisnexis.com/en-us/products/lexis-plus-ai.page}, \\accessed on 25th August 2023} also uses Generative Pre-Trained Transformer (GPT) for generating drafts of documents such as demand letters or client emails. But to the best of our knowledge, there is no tool that helps in drafting 
definitions and terminology in the context of legal documents. 

The paper's main contribution is the \textit{LexDrafter} framework that assists in terminology drafting, i.e., the drafting of \textit{Definitions} articles in legal acts. \textit{LexDrafter} is based on retrieval augmented generation (RAG), where for terms to be defined, relevant text fragments are retrieved and a definition is generated using large language models (LLMs). The advantages of \textit{LexDrafter} are three-fold: first, it aims at harmonizing legal definitions across different legal acts by automating the drafting of \textit{Definitions} articles. Second, human errors are reduced by providing existing term definitions (e.g., for citation) or generating new term definitions, and third, time-consumption is reduced for drafting documents, because users do not have to manually browse through large corpora to ensure consistency of term definitions.

The remainder of this paper is structured as follows. In Section~\ref{sec:relatedwork}, work related to text generation in the legal domain is discussed, followed by a description of the conceptual components underlying the framework in Section~\ref{sec:lexdraftermodel}. Section~\ref{sec:lexdrafterframework} details the two workflows of the \textit{LexDrafter} framework, (1) creation of document and definition corpus, and (2) identification and generation of definitions. In Section~\ref{sec:evaluation}, we present the experiments on generating definitions, and finally, Section~\ref{sec:conclusion} concludes the paper and gives a brief outline of our ongoing work.
\section{Related Work}
\label{sec:relatedwork}
Our work is mostly related to text generation in the legal domain, more specifically EU legislative documents, and there the drafting of \textit{Definitions} articles in respective legislative documents in particular. 

\subsection{Extraction of Legal Definitions}
Using text segmentation and Part-of-Speech (POS) tagging,~\citet{DBLP:conf/iccsa/FernedaPBP12} and~\citet{asi1030022} investigated Brazilian Portuguese texts and Chinese law texts, respectively, to extract legal terms and to build a law ontology. However, manual verification is still required due to a large number of false positives. Rule-based approaches focusing on German laws and cases are used by~\citet{waltl2017automated} for extracting legal definitions and relevant semantic information, such as the year of a dispute. The authors created a taxonomy differentiating between legal definitions, context-extending definitions, and interpretation of legal terms, and further state that legal definition extraction is a difficult task due to additional work required in the field of legal theory.

Addressing the challenge of manually analyzing European regulatory documents, \citet{Damaratskaya2023} focus on semi-automating the analysis by extracting legal definitions and their semantic relations. The author investigates the structure of legal acts and focuses on a single article, i.e., \textit{Definitions} article that specifies legal terms. As punctuation plays an important role in understanding the syntax of legal definitions, using such punctuations, definition terms are extracted, and POS tags are used to extract the explanation of the terms.

\subsection{Document Drafting}
Document creation, document analysis, and document management are the three main branches of legal Document Automation. Different architectures for document generation are reviewed by \citet{DBLP:journals/corr/abs-2109-11603}. For example, Hotdocs\footnote{\url{https://www.hotdocs.com}, \\accessed on 17th August 2023}, DocuPlanner~\cite{DBLP:conf/icail/BrantingLC97}, Virtual Court Action~\cite{Barton1998}, ToXgene~\cite{DBLP:conf/sigmod/BarbosaMKL02}, and FreeMarker\footnote{\url{https://freemarker.apache.org}, \\accessed on 17th August 2023} are template-based tools that allow converting regularly used documents into templates that assist in automated document creation.

The work by \citet{palmirani2011akoma} presents essential elements of the Akoma Ntoso XML standard.
Consuming the Akoma Ntoso format, \citet{DBLP:journals/eis/MarkovicG22} produced legal documents using
a set of legal rules. Furthermore, \citet{DBLP:conf/jurix/PalmiraniG18} used a combination of Akoma Ntoso for marking up legal texts and legal concepts for checking GDPR-compliant public cloud computing services. Supporting the Document Product Line (DPL) methodology, \citet{DBLP:journals/infsof/GomezPCBL14} introduced the DPLFW framework for multi-user, variable content, and reuse-based document generation. \citet{DBLP:journals/eis/MarkovicG22} propose a knowledge-based document assembly by explicitly formulating the legal norms prescribing the content and the form of service contracts. 

In addition to the traditional and knowledge-based document assembly methods discussed, one notable application of RAG proposed by \citet{lewis2020} is document drafting. \citet{markey2024rags} demonstrated the application of RAG in the medical domain to automatically generate documents essential for clinical trials, and \citet{10.1145/3625007.3627505} propose FABULA, a framework that employs RAG to facilitate the report generation process regarding a news event.

Compared to traditional legal NLP tasks, such as classification, information retrieval, and information extraction, currently only a few research is focusing on text generation, i.e., automated drafting of legal documents~\cite{DBLP:journals/corr/abs-2302-12039}. Contract clause generation is a very prominent text generation task, as discussed by \citet{aggarwal-etal-2021-clauserec} and \citet{DBLP:conf/jurix/JoshiBTGV22}. Also, \citet{DBLP:conf/icail/LamCY23} propose combining traditional and generative AI techniques to enhance contract drafting. In their work, the authors investigate the performance of LLMs in contract clause drafting and propose an approach to evaluate the generated clauses by retrieving similar clauses using sentence transformers and then performing content similarity analysis.

Unlike the above approaches, our work focuses on the task of terminology drafting, i.e., \textit{Definitions} articles for EUR-Lex legislative documents. For this, EUR-Lex legal acts are extracted from the EUR-Lex platform, preprocessed, and stored in an Information Retrieval (IR) system. Similar to the approach by \citet{Damaratskaya2023}, existing definitions are extracted from these documents. However, apart from definition extraction itself, citation resolution is additionally carried out for definitions with external references, eventually resulting in so-called definition elements. The stored documents together with definition elements then build the basis for drafting, identifying existing definitions or generating new definitions for selected terms using RAG. 
\section{Document Model and Definition Elements}
\label{sec:lexdraftermodel}

Basis for our \textit{LexDrafter} framework is a corpus of EUR-Lex legal documents. These documents, typically represented in the form of HTML documents, are extracted from the EU platform and preprocessed, following a particular document model. Furthermore, (legal) term definitions in the documents as well as citations related to legal terms are extracted and mapped to definition elements. Both these conceptual components underlying our framework are briefly described in the following. 

\noindent \textbf{Document Model.}
We assume a corpus $D = \{d_1, \ldots, d_n\}$ of (legal) documents. Each document $d_i \in D$ is composed of a sequence of sections $S_i = [s_{i1}, \ldots, s_{ik}]$, with a section providing the coarsest level of (text) granularity. Each section can be decomposed further, e.g., into paragraphs (based on newline elements) or even sentences. Given the natural hierarchical structure of (HTML) documents, the order of elements and their position, respectively, are always preserved and are a property of sections and more fine-granular document components. 

\noindent \textbf{Definition Elements.} For a document $d_i \in D$ and a document section $s_{ij}$, a fragment is simply a sequence of tokens in that section, in most cases a sentence or sequence of sentences in that section. We are in particular interested in fragments that define a legal term, and the process of identifying such fragments is detailed in the next section. Eventually, a set $F_{def} =\{f_1, \ldots, f_l\}$ of definition elements is to be determined for a document collection $D$ such that each definition element $f_i \in F_{def}$ is a tuple $f_i = \langle id_i, t_i, e_i, r_i \rangle$ with
\begin{itemize}
	\item $id_i$ is a unique identifier of the definition element, 
	\item $t_i$ being a term, which is a single word or sequence of words (phrase), that is defined (e.g., ``energy from renewable sources''),
	\item $e_i$ an explanation of the term (i.e., the definition), and 
	\item $r_i$ is a (possibly empty) list of references to definition elements (or rather their $ids$).
\end{itemize}

If an explanation $e_i$ of a term $t_i$ provides a unique explanation, then $r_i$ is empty, otherwise, $r_i$ is a list of respective definition elements ($ids$).
\begin{figure*}[ht]
	\centering
	\vspace{-30pt}
	\includegraphics[width=\textwidth]{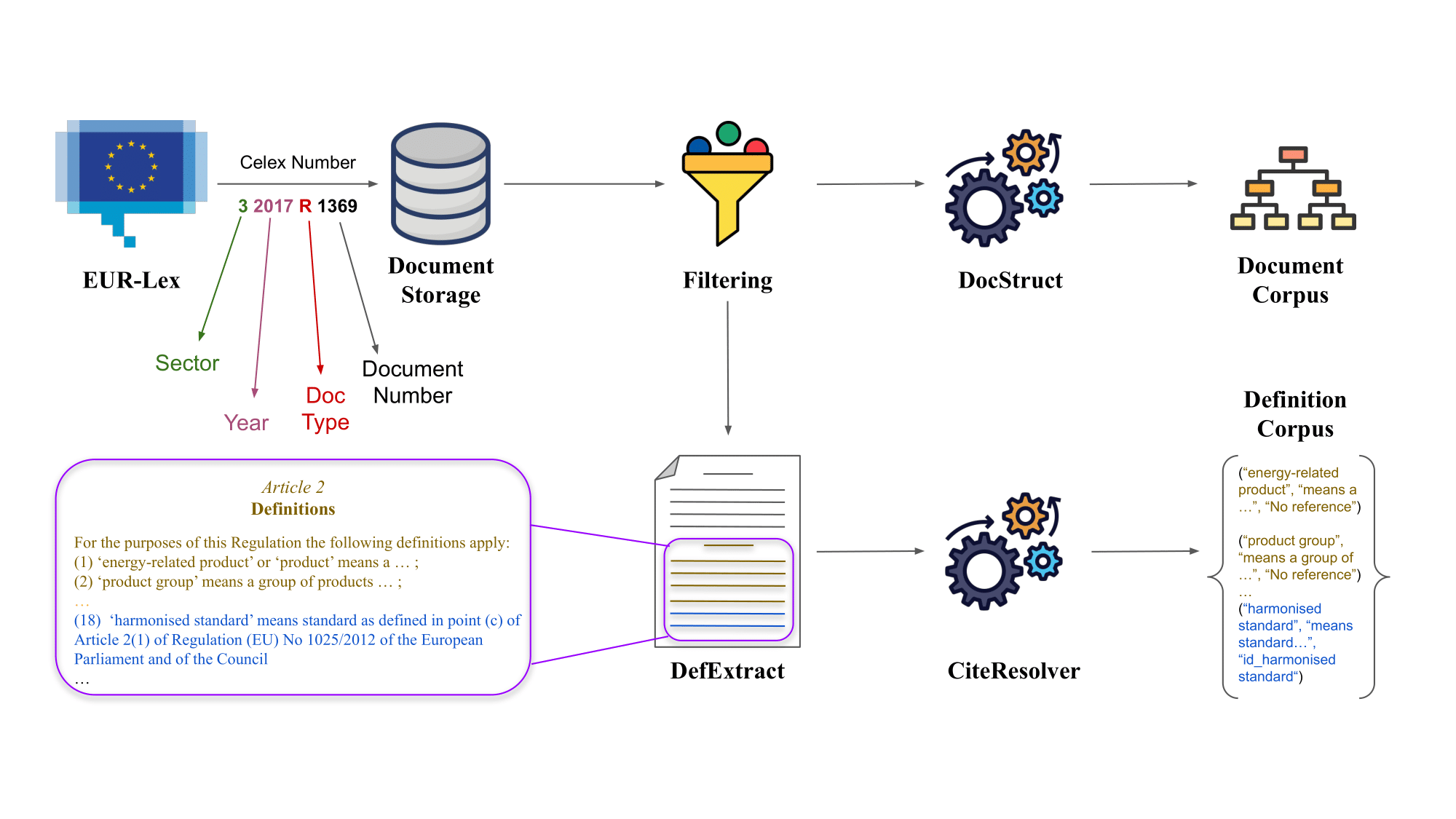}
	\vspace{-35pt}
	\caption{Overview of workflow to build the document and definition corpus, 
		with an example document on ``setting a framework for energy labelling\ldots'' (Celex ID 32017R1369). Legal acts based on Celex ID are extracted from the EUR-Lex platform and filtered to consider only legal acts in HTML format. The \textit{DocStruct} component extracts all information from legal acts and stores it in the document corpus; the \textit{DefExtract} component identifies and extracts definitions from a document. The \textit{CiteResolver} component resolves citations in explanations that have references.}
	\vspace{-10pt}
	\label{fig:lexdrafterworflow1}
\end{figure*}

\section{LexDrafter Framework}
\label{sec:lexdrafterframework}

\textit{LexDrafter} functions help users when drafting a legal act, in particular when drafting document sections has been completed, but the section with terminology definitions (the \textit{Definitions} article) is missing. The \textit{fragments} that include a given \textit{term} in the drafted sections are the key components required by \textit{LexDrafter}, as such fragments provide contextual information for the definition of a term using our RAG approach. Note that automatically identifying terms that need to be defined requires an understanding of legal theory. Therefore, in our work, terms for which definitions need to be determined are selected by the user.

\begin{figure*}[ht]
	\centering
	\vspace{-30pt}
	\includegraphics[width=\textwidth]{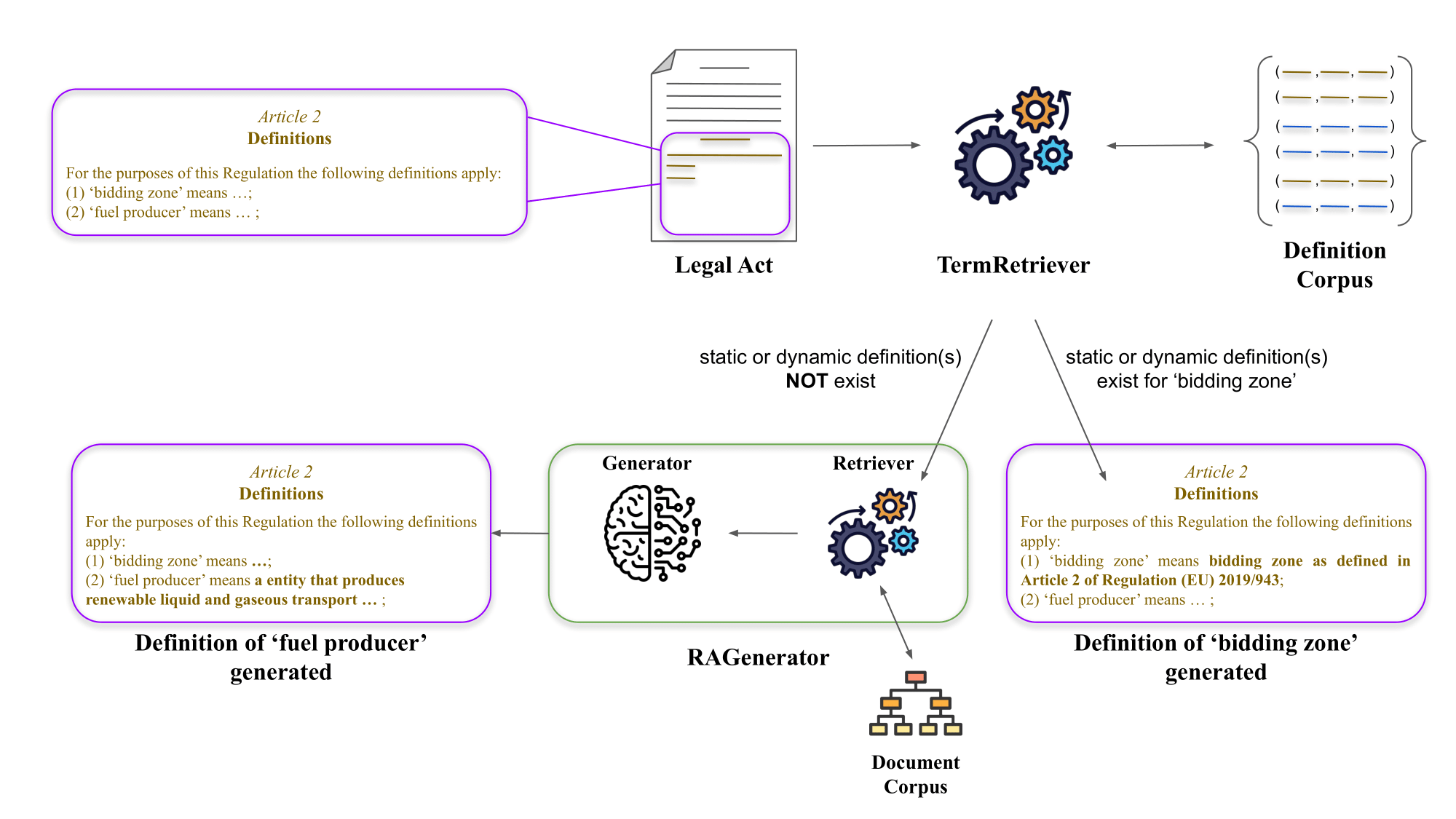}
	\vspace{-15pt}
	\caption{Overview of the Definition Generation workflow with an example document on ``\ldots production of renewable liquid and gaseous transport fuels\ldots'' (Celex ID 32023R1184). Terms to be defined are passed to the \textit{TermRetriever} component to retrieve matching definition elements. `\textit{bidding zone}' is already defined in another legal act; therefore, the definition simply cites this legal act. For `\textit{fuel producer}', a definition does not exist and needs to be generated by the \textit{RAG} component. The \textit{Retriever} subcomponent retrieves fragments and passes these fragments along with the term to be defined to the generator to generate a definition for that term.}
	\vspace{-10pt}
	\label{fig:lexdrafterworkflow2}
\end{figure*}

In the following, we detail the two workflows our \textit{LexDrafter} framework realizes. The first workflow (see also Figure \ref{fig:lexdrafterworflow1}) takes EUR-Lex legal acts (here for the \emph{Energy} domain) as input, preprocesses them, and stores them in an IR system. The second workflow (see Figure \ref{fig:lexdrafterworkflow2}) takes a term selected by the user and either determines existing definitions or generates a definition for that term. The data acquisition process in this work is similar to the one employed by \citet{aumiller-etal-2022-eur}, where a particular legal act web page is crawled to store the text and metadata in an OpenSearch instance.

\subsection{Creating Document and Definition Corpus}
\label{sec:lexdraftermodelcreation}

As illustrated in Figure \ref{fig:lexdrafterworflow1}, from the EUR-Lex website, two data sources are built, (1) the document corpus, which primarily consists of the (properly fragmented) texts of legal acts, including metadata, and (2) term definitions extracted from respective parts of these documents. 

\noindent \textbf{Building Document Corpus.} \citet{chalkidis-etal-2019-extreme} detail the structure of legal documents extracted from the EUR-Lex platform, stating that each legal act is structured into four major zones: \textit{header}, comprising title and name of a legal body enforcing the legal act; \textit{recitals}, consisting of references to legal background of decisions; \textit{main body}, organized as a sequence of articles; and \textit{attachments}, which include appendices and annexes.

In our framework, according to our document model, each header, recital, attachment, and article present in the main body is referred to as a \textit{section}.
The \textit{DocStruct} component shown in Figure~\ref{fig:lexdrafterworflow1} builds the document corpus by extracting respective components from retrieved legal acts, preprocessing and storing them in the IR system.

\noindent \textbf{Building Definition Corpus.} As per JPG~\cite{doi/10.2880/5575}, term consistency must be maintained, and the terms should be defined in a single article in a legal act, called ``\textit{Definitions}'' to avoid wrong interpretations. Therefore, building a definition corpus starts with identifying the \emph{Definitions} article in a document, followed by extracting the definitions, similar to the approach proposed by ~\citet{Damaratskaya2023}. 

Fragments defining a legal term are identified by the \textit{DefExtract} component in Figure \ref{fig:lexdrafterworflow1} using patterns as per ``\textit{Wording Laying Down Definitions}'' (Section C.7 in JPG): 
\begin{itemize}
	\item `\ldots' means \ldots; \textit{[static definition]}
	\item `\ldots' means \ldots as defined in [reference to the static definition]; \textit{[dynamic definition]}
\end{itemize}

Based on this, the legal term within apostrophes and its explanation are extracted. 
For example, consider the definition of the term `\textit{energy from renewable sources}' or `\textit{renewable energy}' present in the document on ``common rules for the internal market for electricity\ldots'' (Celex ID 32019L0944):
\\[0.5ex]
``\textit{`energy from renewable sources'} or \textit{`renewable energy'} means energy from renewable non-fossil sources, namely wind, solar (solar thermal and solar photovoltaic) and geothermal energy, ambient energy, tide, wave and other ocean energy, hydropower, biomass, landfill gas, sewage treatment plant gas, and biogas;'' (Celex ID 32019L0944, Article 2, point 31)

The definition corpus comprises two definition elements for the terms `\textit{energy from renewable sources}' and `\textit{renewable energy}', each with its own id. Both have the same explanation $e$, and as the explanation is unique, the list $r$ of references is empty for both terms.

If the explanation for a term cites other definitions (elements), then the \textit{CiteResolver} component in Figure \ref{fig:lexdrafterworflow1} resolves this citation and populates the references $r$ for that term with the ids of these definition elements. For example, consider the definition of the terms `\textit{energy from renewable sources}' and `\textit{renewable energy}', respectively, in the document on ``the internal market for electricity\ldots'' (Celex ID 32019R0943):

\noindent
``\textit{`energy from renewable sources'} or \textit{`renewable energy'} means energy from renewable sources as defined in point (31) of Article 2 of Directive (EU) 2019/944;'' (Celex ID 32019R0943, Article 2, point 50)

For this example, the \textit{DefExtract} component stores two definition elements in the definition corpus: for the term `\textit{energy from renewable sources}' and for `\textit{renewable energy}'. Both have the explanation ``\textit{means energy from renewable sources as defined in point (31) of Article 2 of Directive (EU) 2019/944}''. Because the explanation cites another definition element (here ``\textit{point (31) of Article 2 of Directive (EU) 2019/944}''), the \textit{CiteResolver} component resolves this citation, determines the identifier of the respective definition element that provides the (static) explanation of the terms `\textit{energy from renewable sources}' and `\textit{renewable energy}', respectively, and stores the identifier in the reference list of both terms.

\subsection{Identifying and Generating Definitions}
\label{subsec:definitiongeneration}

The task of the second workflow shown in Figure \ref{fig:lexdrafterworkflow2} is to identify existing definitions and generate new definitions for terms selected by the user. Existing definitions can easily be identified and retrieved from the definition corpus described above. New definitions are generated using a retrieval augmented generation (RAG) approach. RAG combines pre-trained parametric and non-parametric memory for text generation~\cite{lewis2020}. In the context of \textit{LexDrafter}, the parametric memory is provided by a large language model (LLM), and the non-parametric memory is made up of the fragments and their indexes in the document corpus, respectively. Below we describe (1) how existing term definitions are searched and retrieved, and (2) how a definition for a term is determined using our RAG approach.\\
\noindent \textbf{TermRetriever Component.} If the user would like to know whether for a term in a recently drafted text a definition exists, she/he selects the term, which is then passed to the \textit{TermRetriever} component. The task of this component is to determine if there is a (static) definition for that term, based on the definitions in the legal acts that have been imported into \textit{LexDrafter}. There are three cases: (1) if no definition can be found, the system can generate one (see below); (2) if a single definition is found, its information, e.g., in the form of the definition element is shown to the user. In this case, this information can be used to add a citation to the term, if deemed necessary; (3) if there are multiple definitions, those with the same or similar eurovoc descriptors as the document being drafted are ranked higher when shown to the user. In this case, the user eventually decides what definition to cite in the text for that term, if at all. 

For example, to draft a definition of the term `\textit{bidding zone}' for the document on ``\ldots production of renewable liquid and gaseous transport fuels\ldots'' (Celex ID 32023R1184), the \textit{TermRetriever} would retrieve the definition element for the term `\textit{bidding zone}' that is already defined in the document on ``the internal market for electricity\ldots'' (Celex ID 32019R0943). Thus, the definition of that term could simply be generated as ```\textit{bidding zone}' means bidding zone as defined in Article 2 of Regulation (EU) 2019/943'', without any RAG approach. 

\noindent \textbf{Definition Generation Using RAG.} If the TermRetriever component cannot find a definition for a selected term, the user can request that a definition for the term is generated. The selected term and surrounding text fragments are the two essential parts to generate a term definition. The RAG approach proposed by \citet{lewis2020} and adopted to our framework has two major components, a retriever and a generator. It works as follows: initially, embeddings for the documents (or fragments) are computed and stored, thus building the corpus on which RAG operates. Given a query term and the sections (and fragments) of the document being drafted, the top-$k$ fragments relevant to the query term are determined. 

\textit{LexDrafter} also comprises a retriever and a generator. It is known that a dense embedding retriever often introduces noise by considering document fragments that have terms ``closer'' to the queried term(s). Thus, to reduce such noise in retrieved fragments, we employ a lexical search retriever instead of a dense embedding to retrieve the top-$k$ fragments from the drafted document sections.

As maximum contextual information about a term must be provided to an LLM to generate a term definition, the retrieved fragments are scored based on the term frequency in each fragment, and top-$k$ fragments having the highest term frequency are used. For example, to draft the definition of the term `\textit{fuel producer}' for the document on ``\ldots production of renewable liquid and gaseous transport fuels\ldots'' (Celex ID 32023R1184), from drafted document sections, the retriever determines those fragments with the highest `\textit{fuel producer}' term frequency.

\begin{figure}[!ht]
	\centering
	\vspace{-10pt}
	\includegraphics[width=0.5\textwidth]{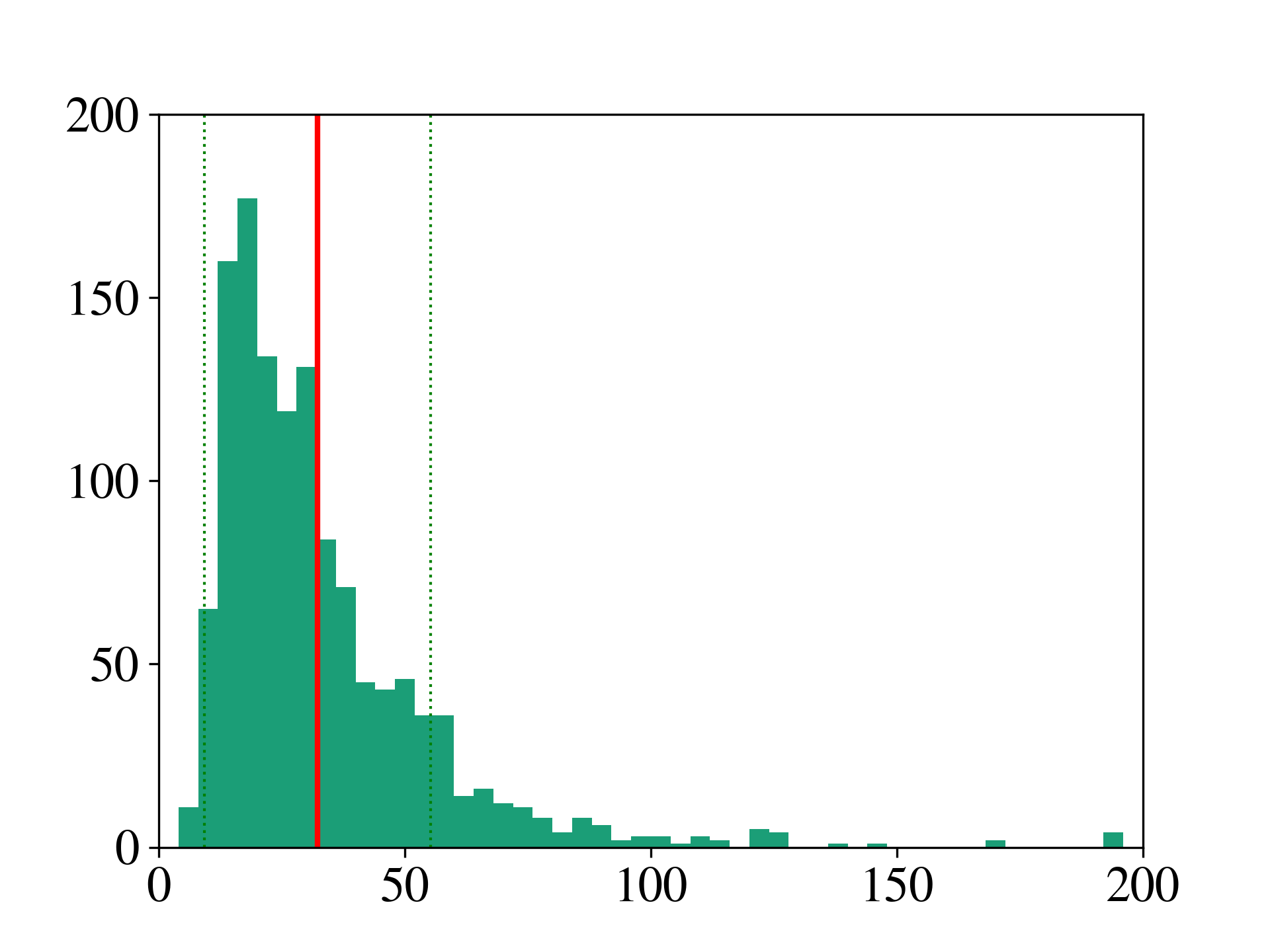}
	\vspace{-3ex}
	\caption{Histogram of the distribution of definition word lengths. Vertical lines show mean length (continuous red) and standard deviation (dotted green lines).}
	\label{fig:tokendistribution}
	\vspace*{-2ex}
\end{figure}

The retrieved fragments and the term to be defined based on the fragments are then passed as a prompt to the generator. The prompt template below is used to generate a definition for the respective term:
\\

\noindent \texttt{Act as a Lawyer drafting European Legislative documents to be published on the Eur-Lex website.}
\\

\noindent \texttt{Define the term: $\{term\}$, based on the sentences provided between the triple dashes where new line characters split different sentences.
	$---
	\{sentences\}
	---$}
\\

\noindent \texttt{Provide a clear and concise definition strictly within 25 to 45 words that accurately conveys the meaning within the context of the sentences.}
\\

\noindent \texttt{Give your output in the following JSON format:
	\\
	$\{\{$\\
	"term": "$\{term\}$"\\
	"definition": "```output text```"\\
	$\}\}$\\
	ONLY return the JSON with the keys: [term, definition], do not add ANYTHING, NO INTERPRETATION!
}
\\

In the prompt, we specify a range of {25 to 45 words}, considering the average number of tokens in the definitions in our corpus, (approximately 32 words), as shown in Figure \ref{fig:tokendistribution}.
\section{Experiments}
\label{sec:evaluation}

For the creation of our document and definition corpus, 551 legal acts from the \textit{Energy} domain between 1949 and 2023 were crawled\footnote{\url{https://eur-lex.europa.eu/search.html?type=named&name=browse-by:legislation-in-force&CC_1_CODED=12}, \\accessed on 22nd March 2024}. The \textit{DocStruct} component shown in Figure~\ref{fig:lexdrafterworflow1} builds the document corpus by extracting respective components from 539 of these legal acts, as only these are in HTML format, preprocesses, and stores them in the IR system. In order to create the definition corpus, those 108 legal acts that contain a \textit{Definitions} article were identified, and a total of 1330 fragments defining 1007 legal terms were detected and stored by the \textit{DefExtract} component as definition elements.

Among the 1007 legal terms, for 787 legal terms, only one definition element exists with static or dynamic definitions, whereas 220 legal terms have multiple definition elements comprising either static or dynamic definitions distributed among 543 definition elements. For example, `\textit{equivalent model}' is often defined as a legal term with 9 definition elements, and surprisingly, all 9 definition elements are static definitions. In order to demonstrate the functionality of \textit{LexDrafter}, the definition elements with static or dynamic definitions for 1007 legal terms are considered. 

\subsection{Large Language Models (LLMs)} 
\label{subsec:llms}

For evaluation, we use Google Colab\footnote{\url{https://colab.google/}, \\accessed on 24th September 2023}, 16GB RAM with V100 GPU, to interact with Vicuna~\cite{DBLP:journals/corr/abs-2306-05685} and LLaMA-2~\cite{DBLP:journals/corr/abs-2307-09288} models. The Hugging Face\footnote{\url{https://huggingface.co/}, \\accessed on 24th September 2023} `text-generation' pipeline is used to interact with the Vicuna-7b-v1.5\footnote{\url{https://huggingface.co/lmsys/vicuna-7b-v1.5}, \\accessed on 24th September 2023} and LLaMA-2-7b\footnote{\url{https://huggingface.co/meta-llama/Llama-2-7b-chat-hf}, \\accessed on 24th September 2023} as a memory-efficient 16-bit quantized model. The important model configuration parameters are as follows: temperature is 0.2, top$\_k$ is 20, top$\_p$ is 0.6, repetition penalty is 1.2, and context length is 4096. For all other model configuration parameters, we use the default value provided by the models. 

\subsection{Results}
\label{subsec:results}
Human-centric metrics, untrained automatic metrics, and machine-learned metrics are the three groups to evaluate text generated with an LLM~\cite{DBLP:journals/corr/abs-2006-14799}. The \textit{LexDrafter} framework is evaluated for the definition generation task, where for a legal term, we investigate the quality of a generated definition with respect to a human-written definition present in the legal acts for the same legal term, i.e., we use ground truth definitions. Therefore, instead of human-centric metrics, untrained automatic metric, i.e., bilingual evaluation understudy (BLEU)~\cite{papineni-etal-2002-bleu} and machine-learned metrics, i.e., BERTScore~\cite{DBLP:conf/iclr/ZhangKWWA20} and bilingual evaluation understudy with representations from transformers (BLEURT)~\cite{sellam-etal-2020-bleurt} are considered to evaluate the quality of generated definitions. 

BLEU measures $n$-gram word overlap of a generated definition with a ground truth definition, whereas BERTScore measures their semantic similarity. BLURT score captures nuances of human language by comparing a generated definition with a ground truth definition on features comprising $n$-gram overlap, semantic similarity, and fluency. Maximizing the BLEURT score signifies fluent, coherent, and semantically accurate generated definitions. In our experiments, open-source LLMs, i.e., Vicuna and LLaMA-2, are used for generating definitions and `distilbert-base-uncased' model in BERTScore and `BLEURT-20'\footnote{\url{https://github.com/google-research/bleurt/blob/master/checkpoints.md}, \\accessed on 20th October 2023} checkpoint in BLEURT is used for evaluation.

\begin{table}[h]
	\setlength{\tabcolsep}{5pt}
	\centering
	\begin{tabular}{|c|c||c||c|}
		\hline \multicolumn{2}{|c||}{\textbf{Metrics} } & \textbf{LLaMA-2} & \textbf{Vicuna} \\
		\hline \multirow{4}{*}{ BLEU } & BLEU-1 & 0.25 & 0.28 \\
		& BLEU-2 & 0.13 & \textbf{0.15} \\
		& BLEU-3 & 0.07 & \textbf{0.09} \\
		& BLEU-4 & 0.04 & 0.05 \\
		\hline \multirow{3}{*}{ BERTScore } & Precision & 0.83 & 0.83 \\
		& Recall & 0.81 & 0.81 \\
		& F1-score & 0.82 & 0.82 \\ 
		\hline \multicolumn{2}{|c||} { BLEURT } & 0.47 & 0.47 \\
		\hline
	\end{tabular}
	\caption{Evaluation of quality of generated definitions. We report $n$-gram BLEU score where $n$ ranges from 1 to 4 to measure the word overlap, BERTScore to measure semantic similarity, and BLEURT measures features comprising $n$-gram, semantic similarity, and fluency. The best results are highlighted in boldface.}
	\label{tab:results}
\end{table}

Table \ref{tab:results} summarizes BLEU, BERTScore, and BLEURT score of generated term definitions using the prompt template from Section~\ref{subsec:definitiongeneration}. The results show that BLEU scores for $n$-gram word overlap where $n$ ranges from 1 to 4 are lower due to different styling and vocabulary used in generated definitions by an LLM when compared to ground truth definitions. On the other hand, BERTScores are higher, demonstrating that generated definitions are semantically similar and have content coverage concerning the term definition present in legal acts. BLEURT scores for both LLMs differ marginally; however, the lower scores are due to training of BLEURT on non-legal domain corpus comprising news articles, blog posts, and wikipedia articles. Thus, a solution to improve the BLEURT score for better insights into the fluency, coherence, and semantics of generated definitions is fine-tuning BLEURT with legal domain corpus. As fine-tuning BLEURT on legal corpus is our ongoing work, fine-tuned BLEURT scores are not presented.

For evaluating the performance of the two LLMs, LLaMA-2 and Vicuna, on the task of generating definitions, our results showed that Vicuna marginally outperformed LLaMA-2 on three metrics: BLEU, BERTScore, and BLEURT. This indicates that Vicuna is better at generating definitions that are fluent, coherent, and semantically similar to the ground truth definitions. A possible explanation for the marginally better performance of Vicuna is the training of Vicuna on a more diverse dataset compared to LLaMA-2, allowing Vicuna to learn a richer representation of the language, enabling it to generate more fluent and coherent definitions. Some examples of generated definitions are shown in Appendix~\ref{appendix:data}.
\section{Conclusions and Ongoing Work}
\label{sec:conclusion}

In this paper, we presented \textit{LexDrafter}, a framework that assists in drafting \textit{Definitions} articles for legislative documents using RAG and existing term definitions present in different legislative documents. 
This work aims to harmonize legal definitions across different legal acts by identifying existing definitions or generating new definition for a term selected by a user, as there are outliers. For example, definitions present in the document on ``\ldots protected geographical indications\ldots'' (Celex ID 32021R0244) do not follow the pattern of having the legal term within apostrophes and document on ``\ldots animal health requirements for movements\ldots'' (Celex ID 32020R0688) defines a term `status free from ``disease''' with ``disease'' as a placeholder.

Our framework realizes two workflows. The first workflow takes EUR-Lex legal acts as input and creates a document corpus and definition corpus. The definition corpus is comprised of extracted term definitions, and the document corpus consists of the (properly fragmented) texts of legal acts. The second workflow takes a term selected by the user as input and either identifies existing definitions or generates a definition for that term using RAG. The RAG approach generates a definition by retrieving fragments using a lexicographic search retriever and passing it to LLMs (Vicuna-7b and LLaMA-7b) as a prompt.

Finally, the functionality of \textit{LexDrafter} is evaluated for a definition generation task with BLEU and BERTScore, where a generated definition is compared to a ground truth definition of a selected term. Table~\ref{tab:results} shows high BERTScore and lower 1 to 4-gram BLEU score, signifying that definitions generated using an LLM show low word overlap but high semantic similarity.

In our ongoing work, we are working to automatically identify terms that need to be defined in a drafted document and consider legal acts from all domains on the EUR-Lex platform to harmonize legal definitions by automating the drafting of \textit{Definitions} articles for cross-domain legal acts.

\section*{Broader Impact \& Ethical Issues}
As the usage of EUR-Lex documents is licensed under CC BY 4.0\footnote{\url{https://eur-lex.europa.eu/content/legal-notice/legal-notice.html}, \\accessed on 20th October 2023} and EUR-Lex documents are used in this work, we do not see any ethical concerns in a release of the code for the \textit{LexDrafter} framework. To our knowledge, legal acts are drafted using the JPG guidelines provided by the European Parliament, the Council and the Commission for persons involved in the drafting of European Union legislation, have undergone review within instances of the European Union, leading to no clear concerns in data quality, especially with respect to potential privacy violations or harmful text content.

\section{Bibliographical References}\label{sec:reference}
\vspace{-22pt}
\bibliographystyle{lrec-coling2024-natbib}
\bibliography{custom}

\begin{thebibliography}{30}
\expandafter\ifx\csname natexlab\endcsname\relax\def\natexlab#1{#1}\fi

\bibitem[{Achachlouei et~al.(2021)Achachlouei, Patil, Joshi, and
  Nair}]{DBLP:journals/corr/abs-2109-11603}
Mohammad~Ahmadi Achachlouei, Omkar Patil, Tarun Joshi, and Vijayan~N. Nair.
  2021.
\newblock \href {https://arxiv.org/abs/2109.11603} {{Document Automation
  Architectures and Technologies: {A} Survey}}.
\newblock \emph{CoRR}, abs/2109.11603.

\bibitem[{Aggarwal et~al.(2021)Aggarwal, Garimella, Srinivasan, N, and
  Jain}]{aggarwal-etal-2021-clauserec}
Vinay Aggarwal, Aparna Garimella, Balaji~Vasan Srinivasan, Anandhavelu N, and
  Rajiv Jain. 2021.
\newblock \href {https://aclanthology.org/2021.emnlp-main.691/}
  {{{C}lause{R}ec: A Clause Recommendation Framework for {AI}-aided Contract
  Authoring}}.
\newblock In \emph{Proceedings of the 2021 Conference on Empirical Methods in
  Natural Language Processing}, pages 8770--8776. Association for Computational
  Linguistics.

\bibitem[{Amaludin et~al.(2021)Amaludin, Wardika, Putra, and
  Paramartha}]{DBLP:conf/jurix/AmaludinWPP21}
Bakhtiar Amaludin, Fitria~Ratna Wardika, Putu Jasprayana~Mudana Putra, and
  I~Gede~Yudi Paramartha. 2021.
\newblock \href {https://ebooks.iospress.nl/volumearticle/58537} {{Analyze the
  Usage of Legal Definitions in Indonesian Regulation Using Text Mining Case
  Study: Treasury and Budget Law}}.
\newblock In \emph{Proceedings of the 34th Conference on Legal Knowledge and
  Information Systems ({JURIX}'21)}, volume 346, pages 107--112. {IOS} Press.

\bibitem[{Aumiller et~al.(2022)Aumiller, Chouhan, and
  Gertz}]{aumiller-etal-2022-eur}
Dennis Aumiller, Ashish Chouhan, and Michael Gertz. 2022.
\newblock \href {https://aclanthology.org/2022.emnlp-main.519/} {{EUR-Lex-Sum:
  A Multi- and Cross-lingual Dataset for Long-form Summarization in the Legal
  Domain}}.
\newblock In \emph{Proceedings of the 2022 Conference on Empirical Methods in
  Natural Language Processing}, pages 7626--7639. Association for Computational
  Linguistics.

\bibitem[{Barbosa et~al.(2002)Barbosa, Mendelzon, Keenleyside, and
  Lyons}]{DBLP:conf/sigmod/BarbosaMKL02}
Denilson Barbosa, Alberto~O. Mendelzon, John Keenleyside, and Kelly~A. Lyons.
  2002.
\newblock \href {https://dl.acm.org/doi/abs/10.1145/564691.564769} {{ToXgene: a
  template-based data generator for {XML}}}.
\newblock In \emph{Proceedings of the 2002 {ACM} {SIGMOD} International
  Conference on Management of Data}, page 616. {ACM}.

\bibitem[{Barton and McKellar(1998)}]{Barton1998}
Karen Barton and Patricia McKellar. 1998.
\newblock \href {https://www.tandfonline.com/doi/abs/10.1080/0968776980060113}
  {The virtual court action: Procedural facilitation in law}.
\newblock \emph{ALT-J: Research in Learning Technology}, 6(1):87--94.

\bibitem[{Branting et~al.(1997)Branting, Lester, and
  Callaway}]{DBLP:conf/icail/BrantingLC97}
Karl Branting, James~C. Lester, and Charles~B. Callaway. 1997.
\newblock \href {https://dl.acm.org/doi/10.1145/261618.261635} {{Automated
  Drafting of Self-Explaining Documents}}.
\newblock In \emph{Proceedings of the Sixth International Conference on
  Artificial Intelligence and Law ({ICAIL}'97)}, pages 72--81. {ACM}.

\bibitem[{Celikyilmaz et~al.(2020)Celikyilmaz, Clark, and
  Gao}]{DBLP:journals/corr/abs-2006-14799}
Asli Celikyilmaz, Elizabeth Clark, and Jianfeng Gao. 2020.
\newblock \href {http://arxiv.org/abs/2006.14799} {{Evaluation of Text
  Generation: {A} Survey}}.
\newblock \emph{CoRR}, abs/2006.14799.

\bibitem[{Chalkidis et~al.(2019)Chalkidis, Fergadiotis, Malakasiotis, Aletras,
  and Androutsopoulos}]{chalkidis-etal-2019-extreme}
Ilias Chalkidis, Emmanouil Fergadiotis, Prodromos Malakasiotis, Nikolaos
  Aletras, and Ion Androutsopoulos. 2019.
\newblock \href {https://aclanthology.org/W19-2209/} {{Extreme Multi-Label
  Legal Text Classification: A Case Study in {EU} Legislation}}.
\newblock In \emph{Proceedings of the Natural Legal Language Processing
  Workshop 2019}, pages 78--87. Association for Computational Linguistics.

\bibitem[{{European Commission and Legal service}(2015)}]{doi/10.2880/5575}
{European Commission and Legal service}. 2015.
\newblock \href {https://data.europa.eu/doi/10.2880/5575} {\emph{{Joint
  practical guide of the European Parliament, the Council and the Commission
  for persons involved in the drafting of European Union legislation}}}.
\newblock Publications Office.

\bibitem[{Ferneda et~al.(2012)Ferneda, do~Prado, Batista, and
  Pinheiro}]{DBLP:conf/iccsa/FernedaPBP12}
Edilson Ferneda, H{\'{e}}rcules~Ant{\^{o}}nio do~Prado, Augusto~Herrmann
  Batista, and Marcello~Sandi Pinheiro. 2012.
\newblock \href
  {https://link.springer.com/chapter/10.1007/978-3-642-31137-6_48} {{Extracting
  Definitions from Brazilian Legal Texts}}.
\newblock In \emph{Proceedings of the 12th International Conference on
  Computational Science and Its Applications ({ICCSA}'12)}, volume 7335, pages
  631--646. Springer.

\bibitem[{Gauci(2016)}]{gauci2016vessel}
Gotthard~Mark Gauci. 2016.
\newblock \href
  {https://heinonline.org/HOL/LandingPage?handle=hein.journals/jmlc47&div=27&id=&page=}
  {{Is it a vessel, a ship or a boat, is it just a craft, or is it merely a
  contrivance}}.
\newblock \emph{Journal of Maritime Law \& Commerce}, 47(4):479--499.

\bibitem[{G{\'{o}}mez et~al.(2014)G{\'{o}}mez, Penad{\'{e}}s, Can{\'{o}}s,
  Borges, and Llavador}]{DBLP:journals/infsof/GomezPCBL14}
Abel G{\'{o}}mez, M.~Carmen Penad{\'{e}}s, Jos{\'{e}}~H. Can{\'{o}}s, Marcos
  R.~S. Borges, and Manuel Llavador. 2014.
\newblock \href
  {https://www.sciencedirect.com/science/article/pii/S0950584913002358} {A
  framework for variable content document generation with multiple actors}.
\newblock \emph{Information and Software Technology}, 56(9):1101--1121.

\bibitem[{Hwang et~al.(2018)Hwang, Hsueh, and Chang}]{asi1030022}
Ren-Hung Hwang, Yu-Ling Hsueh, and Yu-Ting Chang. 2018.
\newblock \href {https://doi.org/10.3390/asi1030022} {{Building a Taiwan Law
  Ontology Based on Automatic Legal Definition Extraction}}.
\newblock \emph{Applied System Innovation}, 1(3):1--20.

\bibitem[{Joshi et~al.(2022)Joshi, Balaji, Thomas, Garimella, and
  Varma}]{DBLP:conf/jurix/JoshiBTGV22}
Sagar Joshi, Sumanth Balaji, Jerrin Thomas, Aparna Garimella, and Vasudeva
  Varma. 2022.
\newblock \href {https://ebooks.iospress.nl/doi/10.3233/FAIA220450}
  {{Investigating Strategies for Clause Recommendation}}.
\newblock In \emph{Proceedings of the 35th Legal Knowledge and Information
  Systems ({JURIX}'22)}, volume 362, pages 73--82. {IOS} Press.

\bibitem[{Katz et~al.(2023)Katz, Hartung, Gerlach, Jana, and
  II}]{DBLP:journals/corr/abs-2302-12039}
Daniel~Martin Katz, Dirk Hartung, Lauritz Gerlach, Abhik Jana, and Michael
  J.~Bommarito II. 2023.
\newblock \href {https://arxiv.org/abs/2302.12039} {{Natural Language
  Processing in the Legal Domain}}.
\newblock \emph{CoRR}, abs/2302.12039.

\bibitem[{Lam et~al.(2023)Lam, Cheng, and Yeong}]{DBLP:conf/icail/LamCY23}
Kwok{-}Yan Lam, Victor C.~W. Cheng, and Zee~Kin Yeong. 2023.
\newblock \href {https://ceur-ws.org/Vol-3423/paper7.pdf} {{Applying Large
  Language Models for Enhancing Contract Drafting}}.
\newblock In \emph{Proceedings of the Third International Workshop on
  Artificial Intelligence and Intelligent Assistance for Legal Professionals in
  the Digital Workplace (LegalAIIA 2023)}, volume 3423, pages 70--80. CEUR-WS.

\bibitem[{Lewis et~al.(2020)Lewis, Perez, Piktus, Petroni, Karpukhin, Goyal,
  K\"{u}ttler, Lewis, Yih, Rockt\"{a}schel, Riedel, and Kiela}]{lewis2020}
Patrick Lewis, Ethan Perez, Aleksandra Piktus, Fabio Petroni, Vladimir
  Karpukhin, Naman Goyal, Heinrich K\"{u}ttler, Mike Lewis, Wen-tau Yih, Tim
  Rockt\"{a}schel, Sebastian Riedel, and Douwe Kiela. 2020.
\newblock \href {https://dl.acm.org/doi/abs/10.5555/3495724.3496517}
  {{Retrieval-Augmented Generation for Knowledge-Intensive NLP Tasks}}.
\newblock In \emph{Proceedings of the 34th International Conference on Neural
  Information Processing Systems ({NIPS}'20)}, volume 793, pages 9459–--9474.
  {ACM}.

\bibitem[{Markey et~al.(2024)Markey, El-Mansouri, Rensonnet, van Langen, and
  Meier}]{markey2024rags}
Nigel Markey, Ilyass El-Mansouri, Gaetan Rensonnet, Casper van Langen, and
  Christoph Meier. 2024.
\newblock \href {https://arxiv.org/abs/2402.16406} {{From RAGs to riches: Using
  large language models to write documents for clinical trials}}.
\newblock \emph{CoRR}, abs/2402.16406.

\bibitem[{Marković and Gostojić(2022)}]{DBLP:journals/eis/MarkovicG22}
Marko Marković and Stevan Gostojić. 2022.
\newblock \href {https://doi.org/10.1080/17517575.2020.1793389} {A
  knowledge-based document assembly method to support semantic interoperability
  of enterprise information systems}.
\newblock \emph{Enterprise Information Systems}, 16(5).

\bibitem[{Palmirani and Governatori(2018)}]{DBLP:conf/jurix/PalmiraniG18}
Monica Palmirani and Guido Governatori. 2018.
\newblock \href {https://ebooks.iospress.nl/volumearticle/50839} {{Modelling
  Legal Knowledge for {GDPR} Compliance Checking}}.
\newblock In \emph{Proceedings of the 31st Conference on Legal Knowledge and
  Information Systems ({JURIX}'18)}, volume 313, pages 101--110. {IOS} Press.

\bibitem[{Palmirani and Vitali(2011)}]{palmirani2011akoma}
Monica Palmirani and Fabio Vitali. 2011.
\newblock \href {https://link.springer.com/chapter/10.1007/978-94-007-1887-6_6}
  {{Akoma-Ntoso for Legal Documents}}.
\newblock In \emph{Legislative {{XML}} for the Semantic Web: {{Principles}},
  Models, Standards for Document Management}, pages 75--100. {Springer}.

\bibitem[{Papineni et~al.(2002)Papineni, Roukos, Ward, and
  Zhu}]{papineni-etal-2002-bleu}
Kishore Papineni, Salim Roukos, Todd Ward, and Wei-Jing Zhu. 2002.
\newblock \href {https://aclanthology.org/P02-1040/} {{{B}LEU: a Method for
  Automatic Evaluation of Machine Translation}}.
\newblock In \emph{Proceedings of the 40th Annual Meeting of the Association
  for Computational Linguistics}, pages 311--318. Association for Computational
  Linguistics.

\bibitem[{Ranade and Joshi(2024)}]{10.1145/3625007.3627505}
Priyanka Ranade and Anupam Joshi. 2024.
\newblock \href {https://dl.acm.org/doi/10.1145/3625007.3627505} {{FABULA:
  Intelligence Report Generation Using Retrieval-Augmented Narrative
  Construction}}.
\newblock In \emph{Proceedings of the 2023 IEEE/ACM International Conference on
  Advances in Social Networks Analysis and Mining (ASONAM'23)}, pages 603--610.
  Association for Computing Machinery.

\bibitem[{Sai et~al.({2023})Sai, Damaratskaya, Winter, and
  {Rinderle-Ma}}]{Damaratskaya2023}
Catherine Sai, Anastasiya Damaratskaya, Karolin Winter, and Stefanie
  {Rinderle-Ma}. {2023}.
\newblock \href
  {https://link.springer.com/chapter/10.1007/978-3-031-47112-4_14}
  {{Identification and Visualization of Legal Definitions and Legal Term
  Relations}}.
\newblock In \emph{{Advances in Conceptual Modeling}}, pages {151--161}.
  {Springer Nature Switzerland}.

\bibitem[{Sellam et~al.(2020)Sellam, Das, and Parikh}]{sellam-etal-2020-bleurt}
Thibault Sellam, Dipanjan Das, and Ankur Parikh. 2020.
\newblock \href {https://doi.org/10.18653/v1/2020.acl-main.704} {{{BLEURT}:
  Learning Robust Metrics for Text Generation}}.
\newblock In \emph{Proceedings of the 58th Annual Meeting of the Association
  for Computational Linguistics}, pages 7881--7892, Online. Association for
  Computational Linguistics.

\bibitem[{Touvron et~al.(2023)Touvron, Martin, Stone, Albert, Almahairi,
  Babaei, Bashlykov, \ldots, Zhang, Fan, Kambadur, Narang, Rodriguez, Stojnic,
  Edunov, and Scialom}]{DBLP:journals/corr/abs-2307-09288}
Hugo Touvron, Louis Martin, Kevin Stone, Peter Albert, Amjad Almahairi, Yasmine
  Babaei, Nikolay Bashlykov, \ldots, Yuchen Zhang, Angela Fan, Melanie
  Kambadur, Sharan Narang, Aur{\'{e}}lien Rodriguez, Robert Stojnic, Sergey
  Edunov, and Thomas Scialom. 2023.
\newblock \href {https://arxiv.org/abs/2307.09288} {{Llama 2: Open Foundation
  and Fine-Tuned Chat Models}}.
\newblock \emph{CoRR}, abs/2307.09288.

\bibitem[{Waltl et~al.(2017)Waltl, Landthaler, Scepankova, Matthes, Geiger,
  Stocker, and Schneider}]{waltl2017automated}
Bernhard Waltl, J{\"o}rg Landthaler, Elena Scepankova, Florian Matthes, THOMAS
  Geiger, Christoph Stocker, and Christian Schneider. 2017.
\newblock \href
  {https://wwwmatthes.in.tum.de/pages/xsepf8e1admj/Automated-extraction-of-semantic-information-from-german-legal-documents}
  {Automated extraction of semantic information from german legal documents}.
\newblock In \emph{IRIS: Internationales Rechtsinformatik Symposium, Austria}.

\bibitem[{Zhang et~al.(2020)Zhang, Kishore, Wu, Weinberger, and
  Artzi}]{DBLP:conf/iclr/ZhangKWWA20}
Tianyi Zhang, Varsha Kishore, Felix Wu, Kilian~Q. Weinberger, and Yoav Artzi.
  2020.
\newblock \href {https://arxiv.org/abs/1904.09675} {{BERTScore: Evaluating Text
  Generation with {BERT}}}.
\newblock In \emph{Proceedings of the Eighth International Conference on
  Learning Representations ({ICLR}'20)}. OpenReview.net.

\bibitem[{Zheng et~al.(2023)Zheng, Chiang, Sheng, Zhuang, Wu, Zhuang, Lin, Li,
  Li, Xing, Zhang, Gonzalez, and Stoica}]{DBLP:journals/corr/abs-2306-05685}
Lianmin Zheng, Wei{-}Lin Chiang, Ying Sheng, Siyuan Zhuang, Zhanghao Wu,
  Yonghao Zhuang, Zi~Lin, Zhuohan Li, Dacheng Li, Eric~P. Xing, Hao Zhang,
  Joseph~E. Gonzalez, and Ion Stoica. 2023.
\newblock \href {https://arxiv.org/abs/2306.05685} {{Judging LLM-as-a-judge
  with MT-Bench and Chatbot Arena}}.
\newblock \emph{CoRR}, abs/2306.05685.

\end{thebibliography}


\appendix
\section{Examples}
\label{appendix:data}
Below we show some examples of definitions generated by the RAG approach with LLaMA-2 for a term present in a legal act. \textit{original definition} and \textit{generated definition} state definition present in the legal act, and generated definition respectively.
\\

\noindent `\textbf{legal term}': `rules'\\
\noindent `\textbf{Celex ID}': `32021Q0618(02)'\\
\noindent `\textbf{original definition}': ```rules' means these Rules of the Supply Agency of the European Atomic Energy Community determining the manner in which demand is to be balanced against the supply of ores, source materials and special fissile materials;''\\
\noindent `\textbf{generated definition}': ```rules' means procedures and guidelines established by the european agency for the supply of nuclear materials for the purpose of ensuring the regular and equitable supply of nuclear materials to community users;''
\\

\noindent `\textbf{legal term}': `generic ecodesign requirement'\\
\noindent `\textbf{Celex ID}': `32009L0125'\\
\noindent `\textbf{original definition}': ```generic ecodesign requirement' means any ecodesign requirement based on the ecological profile as a whole of a product without set limit values for particular environmental aspects;'\\
\noindent `\textbf{generated definition}': ```generic ecodesign requirement' means a method for improving environmental performance of products without setting limit values, focusing on significant environmental aspects;''
\\

\noindent `\textbf{legal term}': `refrigerating appliance'\\
\noindent `\textbf{Celex ID}': `32019R2016'\\
\noindent `\textbf{original definition}': ```refrigerating appliance' means an insulated cabinet with one or more compartments that are controlled at specific temperatures, cooled by natural or forced convection whereby the cooling is obtained by one or more energy consuming means;'\\
\noindent `\textbf{generated definition}': ```refrigerating appliance' means a device that cools or freezes matter by transferring heat from the interior of the appliance to the exterior, using vapor compression or other means, and may include one or more compartments for storing food or drinks;''
\\

\noindent `\textbf{legal term}': `abandoned land'\\
\noindent `\textbf{Celex ID}': `32019R0807'\\
\noindent `\textbf{original definition}': ```abandoned land' means unused land, which was used in the past for the cultivation of food and feed crops but where the cultivation of food and feed crops was stopped due to biophysical or socioeconomic constraints;'\\
\noindent `\textbf{generated definition}': ```abandoned land' means land that has been unused or neglected for an extended period of time, often due to economic, environmental, or social reasons;''
\\

\noindent `\textbf{legal term}': `adequacy'\\
\noindent `\textbf{Celex ID}': `32017R1485'\\
\noindent `\textbf{original definition}': ```adequacy' means the ability of in-feeds into an area to meet the load in that area;'\\
\noindent `\textbf{generated definition}': ```adequacy' means the absence of a mismatch between the available supply of electricity in a control area and the demand for electricity in that area, as determined through a control area adequacy analysis;''
\\

\noindent `\textbf{legal term}': `aggregates'\\
\noindent `\textbf{Celex ID}': `32008R1099'\\
\noindent `\textbf{original definition}': ```aggregates' means data aggregated at national level on the treatment or use of energy products, namely production, trade, stocks, transformation, consumption, and structural characteristics of the energy system such as installed capacities for electricity generation or production capacities for oil products;'\\
\noindent `\textbf{generated definition}': ```aggregates' means quantities of energy products used as raw materials in different sectors and not consumed as a fuel or transformed into another fuel, including quantities declared for electricity and heat production, gross and net calorific values, and consumption in main activity producer plants and energy end-use specification;''
\end{document}